\documentclass[runningheads]{llncs}
\usepackage{graphicx}
\usepackage{amsmath,amssymb} % define this before the line numbering.
\usepackage{color}

\usepackage{multirow}
\usepackage{mathrsfs}
\usepackage{enumitem}
\usepackage{makecell}
\usepackage{tabulary}
\usepackage{pifont}
\usepackage{subfigure}
\usepackage[colorlinks,colorlinks=true,linkcolor=black,urlcolor=black,anchorcolor=black,citecolor=black]{hyperref}

\begin{document}
\title{Exploring Visual Relationship \\for Image Captioning}
% Replace with your title

\titlerunning{Exploring Visual Relationship for Image Captioning}
% Replace with a meaningful short version of your title
%
\author{Ting Yao\inst{1} \and
Yingwei Pan\inst{1} \and
Yehao Li\inst{2} \and
Tao Mei\inst{1}
}
%
%Please write out author names in full in the paper, i.e. full given and family names.
%If any authors have names that can be parsed into FirstName LastName in multiple ways, please include the correct parsing, in a comment to the volume editors:
%\index{Lastnames, Firstnames}
%(Do not uncomment it, because you may introduce extra index items if you do that, we will use scripts for introducing index entries...)
\authorrunning{T. Yao, Y. Pan, Y. Li, and T. Mei}
% Replace with shorter version of the author list. If there are more authors than fits a line, please use A. Author et al.
%

\institute{JD AI Research, Beijing, China \and
Sun Yat-sen University, Guangzhou, China\\
\email{\{tingyao.ustc,panyw.ustc,yehaoli.sysu\}@gmail.com, tmei@live.com}}
\maketitle              % typeset the header of the contribution
\begin{abstract}

It is always well believed that modeling relationships between objects would be helpful for representing and eventually describing an image. Nevertheless, there has not been evidence in support of the idea on image description generation. In this paper, we introduce a new design to explore the connections between objects for image captioning under the umbrella of attention-based encoder-decoder framework. Specifically, we present Graph Convolutional Networks plus Long Short-Term Memory (dubbed as GCN-LSTM) architecture that novelly integrates both semantic and spatial object relationships into image encoder. Technically, we build graphs over the detected objects in an image based on their spatial and semantic connections. The representations of each region proposed on objects are then refined by leveraging graph structure through GCN. With the learnt region-level features, our GCN-LSTM capitalizes on LSTM-based captioning framework with attention mechanism for sentence generation. Extensive experiments are conducted on COCO image captioning dataset, and superior results are reported when comparing to state-of-the-art approaches. More remarkably, GCN-LSTM increases CIDEr-D performance from 120.1\% to 128.7\% on COCO testing set.

\end{abstract}
\section{Introduction}
\begin{figure*}[!tb]
\centering {\includegraphics[width=1\textwidth]{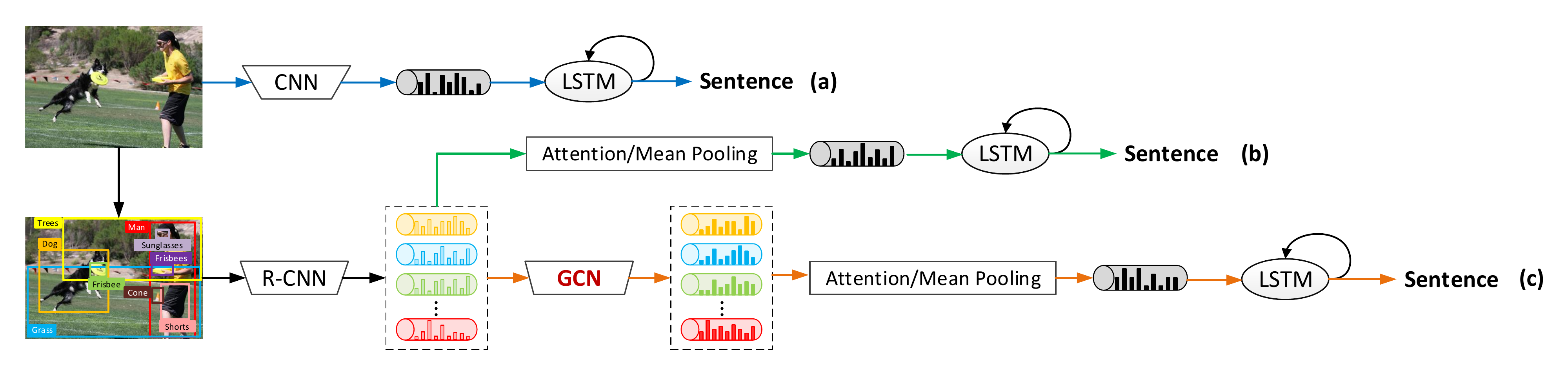}}
\caption{Visual representations generated by image encoder in (a) CNN plus LSTM, (b) R-CNN plus LSTM, and (c) our GCN-LSTM for image captioning.}
\label{fig:fig1}
\end{figure*}

The recent advances in deep neural networks have convincingly demonstrated high capability in learning vision models particularly for recognition. The achievements make a further step towards the ultimate goal of image understanding, which is to automatically describe image content with a complete and natural sentence or referred to as image captioning problem. The typical solutions \cite{Donahue14,Vinyals14,Xu:ICML15,yao2017boosting} of image captioning are inspired by machine translation and equivalent to translating an image to a text. As illustrated in Figure \ref{fig:fig1} (a) and (b), a Convolutional Neural Network (CNN) or Region-based CNN (R-CNN) is usually exploited to encode an image and a decoder of Recurrent Neural Network (RNN) w/ or w/o attention mechanism is utilized to generate the sentence, one word at each time step. Regardless of these different versions of CNN plus RNN image captioning framework, a common issue not fully studied is how visual relationships should be leveraged in view that the mutual correlations or interactions between objects are the natural basis for describing an image.

Visual relationships characterize the interactions or relative positions between objects detected in an image. The detection of visual relationships involves not only localizing and recognizing objects, but also classifying the interaction (predicate) between each pair of objects. In general, the relationship can be represented as $\langle$\emph{subject-predicate-object}$\rangle$, e.g., $\langle$\emph{man-eating-sandwich}$\rangle$ or $\langle$\emph{dog-inside-car}$\rangle$. In the literature, it is well recognized that reasoning such visual relationships is crucial to a richer semantic understanding \cite{Yikang:ICCV17,lu2016visual} of the visual world. Nevertheless, the fact that the objects could be with a wide range of scales, at arbitrary positions in an image and from different categories results in difficulty in determining the type of relationships. In this paper, we take the advantages of the inherent relationships between objects for interpreting the images holistically and novelly explore the use of visual connections to enhance image encoder for image captioning. Our basic design is to model the relationships on both semantic and spatial levels, and integrate the connections into image encoder to produce relation-aware region-level representations. As a result, we endow image representations with more power when feeding into sentence decoder.

By consolidating the idea of modeling visual relationship for image captioning, we present a novel Graph Convolutional Networks plus Long Short-Term Memory (GCN-LSTM) architecture, as conceptually shown in Figure \ref{fig:fig1} (c). Specifically, Faster R-CNN is firstly implemented to propose a set of salient image regions. We build semantic graph with directed edges on the detected regions, where the vertex represents each region and the edge denotes the relationship (predicate) between each pair of regions which is predicted by semantic relationship detector learnt on Visual Genome \cite{krishna2017visual}. Similarly, spatial graph is also constructed on the regions and the edge between regions models relative geometrical relationship. Graph Convolutional Networks are then exploited to enrich region representations with visual relationship in the structured semantic and spatial graph respectively. After that, the learnt relation-aware region representations on each kind of relationships are feed into one individual attention LSTM decoder to generate the sentence. In the inference stage, to fuse the outputs of two decoders, we linearly average the predicted score distributions on words from two decoders at each time step and pop out the word with the highest probability as the input word to both decoders at the next step.

The main contribution of this work is the proposal of the use of visual relationship for enriching region-level representations and eventually enhancing image captioning. This also leads to the elegant views of what kind of visual relationships could be built between objects, and how to nicely leverage such visual relationships to learn more informative and relation-aware region representations for image captioning, which are problems not yet fully understood.

\section{Related Work}\label{sec:RW}
\textbf{Image Captioning.} With the prevalence of deep learning \cite{Alex:NIPS12} in computer vision, the dominant paradigm in modern image captioning is sequence learning methods \cite{Donahue14,Vinyals14,Xu:ICML15,yao2017novel,yao2017boosting,You:CVPR16} which utilize CNN plus RNN model to generate novel sentences with flexible syntactical structures. For instance, Vinyals \emph{et al.} propose an end-to-end neural networks architecture by utilizing LSTM to generate sentence for an image in \cite{Vinyals14}, which is further incorporated with soft/hard attention mechanism in \cite{Xu:ICML15} to automatically focus on salient objects when generating corresponding words. Instead of activating visual attention over image for every generated word, \cite{Xiong2016MetaMind} develops an adaptive attention encoder-decoder model for automatically deciding when to rely on visual signals/language model. Recently, in \cite{Wu:CVPR16,yao2017boosting}, semantic attributes are shown to clearly boost image captioning when injected into CNN plus RNN model and such attributes can be further leveraged as semantic attention \cite{You:CVPR16} to enhance image captioning. Most recently, a novel attention based encoder-decoder model \cite{anderson2017bottom} is proposed to detect a set of salient image regions via bottom-up attention mechanism and then attend to the salient regions with top-down attention mechanism for sentence generation.

\textbf{Visual Relationship Detection.} Research on visual relationship detection has attracted increasing attention. Some early works \cite{galleguillos2008object,gould2008multi} attempt to learn four spatial relations (i.e., ``above", ``below", ``inside" and ``around") to improve segmentation. Later on, semantic relations (e.g., actions or interactions) between objects are explored in \cite{divvala2014learning,sadeghi2011recognition} where each possible combination of semantic relation is taken as a visual phrase class and the visual relationship detection is formulated as a classification task. Recently, quite a few works \cite{dai2017detecting,Yikang:ICCV17,lu2016visual,plummer2017phrase,xu2017scene} design deep learning based architectures for visual relationship detection. \cite{xu2017scene} treats visual relationship as the directed edges to connect two object nodes in the scene graph and the relationships are inferred along the processing of constructing scene graph in an iterative way. \cite{dai2017detecting,Yikang:ICCV17} directly learn the visual features for relationship prediction based on additional union bounding boxes which cover object and subject together. In \cite{lu2016visual,plummer2017phrase}, the linguistic cues of the participating objects/captions are further considered for visual relationship detection.

\textbf{Summary.} In short, our approach in this paper belongs to sequence learning method for image captioning. Similar to previous approaches \cite{anderson2017bottom,fu2017aligning}, GCN-LSTM explores visual attention over the detected image regions of objects for sentence generation. The novelty is on the exploitation of semantic and spatial relations between objects for image captioning, that has not been previously explored. In particular, both of the two kinds of visual relationships are seamlessly integrated into LSTM-based captioning framework via GCN, targeting for producing relation-aware region representations and thus potentially enhancing the quality of generated sentence through emphasizing the object relations.

\section{Image Captioning by Exploring Visual Relationship}
\begin{figure*}[!tb]
\centering {\includegraphics[width=0.99\textwidth]{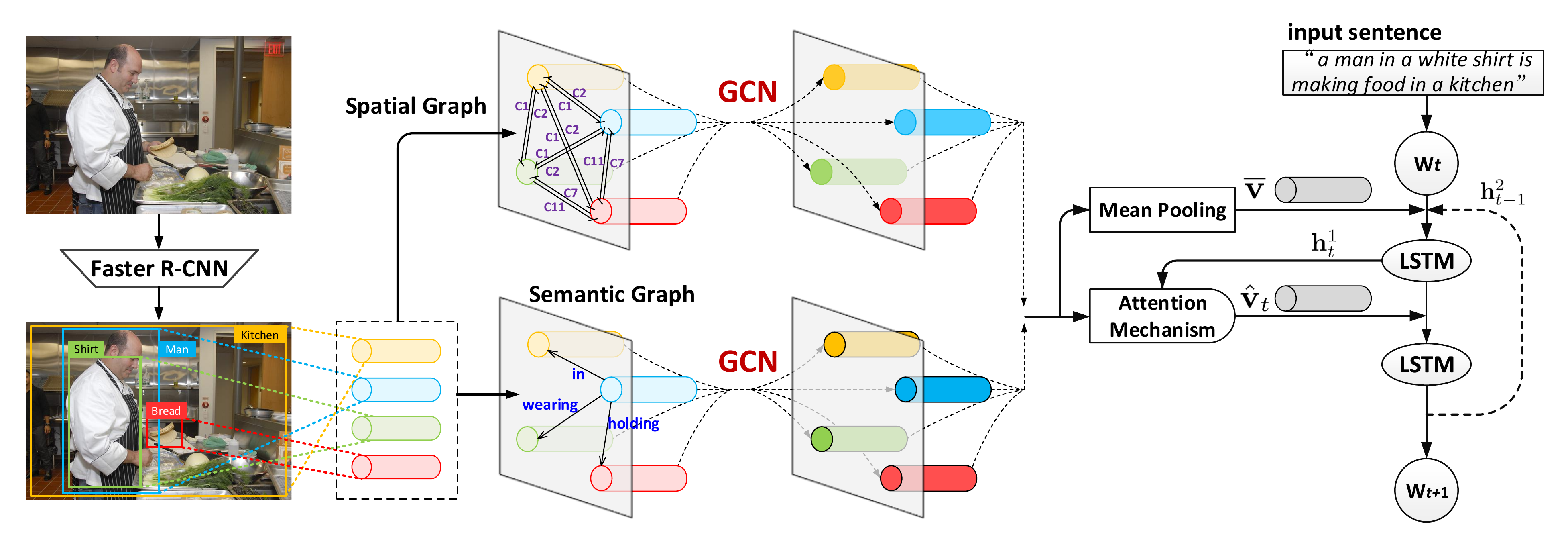}}
\caption{An overview of our Graph Convolutional Networks plus Long Short-Term Memory (GCN-LSTM) for image captioning (better viewed in color). Faster R-CNN is first leveraged to detect a set of salient image regions. Next, semantic/spatial graph is built with directional edges on the detected regions, where the vertex represents each region and the edge denotes the semantic/spatial relationship in between. Graph Convolutional Networks (GCN) is then exploited to contextually encode regions with visual relationship in the structured semantic/spatial graph. After that, the learnt relation-aware region-level features from each kind of graph are feed into one individual attention LSTM decoder for sentence generation. In the inference stage, we adopt a late fusion scheme to linearly fuse the results from two decoders.}
\label{fig:fig2}
\end{figure*}

We devise our Graph Convolutional Networks plus Long Short-Term Memory (GCN-LSTM) architecture to generate image descriptions by additionally incorporating both semantic and spatial object relationships. GCN-LSTM firstly utilizes an object detection module (e.g., Faster R-CNN \cite{ren2015faster}) to detect objects within images, aiming for encoding and generalizing the whole image into a set of salient image regions containing objects. Semantic and spatial relation graphs are then constructed over all the detected image regions of objects based on their semantic and spatial connections, respectively. Next, the training of GCN-LSTM is performed by contextually encoding the whole image region set with semantic or spatial graph structure via GCN, resulting in relation-aware region representations. All of encoded relation-aware region representations are further injected into LSTM-based captioning framework, enabling region-level attention mechanism for sentence generation. An overview of our image captioning architecture is illustrated in Figure \ref{fig:fig2}.

\subsection{Problem Formulation}
Suppose we have an image ${I}$ to be described by a textual sentence $\mathcal {S}$, where $\mathcal{S} = \{w_1, w_2, ..., w_{N_s}\}$ consisting of $N_s$ words. Let ${\bf{w}}_t\in {{\mathbb{R}}^{D_s}}$ denote the $D_s$-dimensional textual feature of the $t$-th word in sentence $\mathcal{S}$. Faster R-CNN is firstly leveraged to produce the set of detected objects $\mathcal{V}=\{{v_i}\}^{K}_{i=1}$ with $K$ image regions of objects in ${I}$ and ${{\bf{v}}_i}\in {{\mathbb{R}}^{D_v}}$ denotes the $D_v$-dimensional feature of each image region. Furthermore, by treating each image region $v_i$ as one vertex, we can construct semantic graph $\mathcal{G}_{sem}=(\mathcal{V},\mathcal{E}_{sem})$ and spatial graph $\mathcal{G}_{spa}=(\mathcal{V},\mathcal{E}_{spa})$, where $\mathcal{E}_{sem}$ and $\mathcal{E}_{spa}$ denotes the set of semantic and spatial relation edges between region vertices, respectively. More details about how we mine the visual relationships between objects and construct the semantic and spatial graphs will be elaborated in Section \ref{sec:SG}.

Inspired by the recent successes of sequence models leveraged in image/video captioning \cite{pan2016jointly,pan2017video,Vinyals14} and region-level attention mechanism \cite{anderson2017bottom,fu2017aligning}, we aim to formulate our image captioning model in a R-CNN plus RNN scheme. Our R-CNN plus RNN method firstly interprets the given image as a set of image regions with R-CNN, then uniquely encodes them into relation-aware features conditioned on semantic/spatial graph, and finally decodes them to each target output word via attention LSTM decoder. Derived from the idea of Graph Convolutional Networks \cite{kipf2017semi,marcheggiani2017encoding}, we leverage a GCN module in image encoder to contextually refine the representation of each image region, which is endowed with the inherent visual relationships between objects. Hence, the sentence generation problem we explore here can be formulated by minimizing the following energy loss function:
\begin{equation}\label{Eq:Eq1}
E(\mathcal{V}, \mathcal{G}, {\mathcal {S}}) = -\log {\Pr{({\mathcal {S}}|\mathcal{V}, \mathcal{G})}},
\end{equation}
which is the negative log probability of the correct textual sentence given the detected image regions of objects $\mathcal{V}$ and constructed relation graph $\mathcal{G}$. Note that we use $\mathcal{G} \in \{\mathcal{G}_{sem}, \mathcal{G}_{spa}\} $ for simplicity, i.e., $\mathcal{G}$ denotes either semantic graph $\mathcal{G}_{sem}$ or spatial graph $\mathcal{G}_{spa}$. Here the negative $\log$ probability is typically measured with cross entropy loss, which inevitably results in the discrepancy of evaluation between training and inference. Accordingly, to further boost our captioning model by amending such discrepancy, we can directly optimize the LSTM with expected sentence-level reward loss as in \cite{li2018jointly,Liu:2016PGSPIDEr,rennie2017self}.

\begin{figure*}[!tb]
\centering {\includegraphics[width=0.77\textwidth]{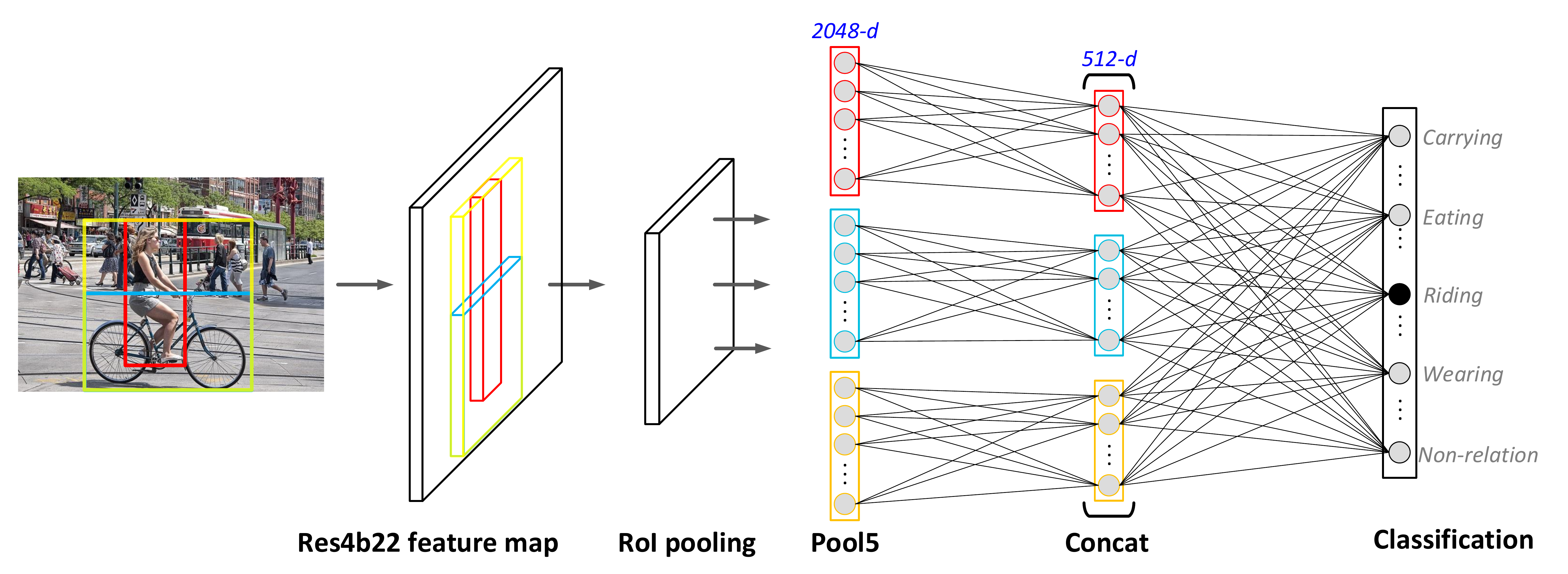}}
\caption{Detection model for semantic relation $\langle$\emph{subject-predicate-object}$\rangle$ (red: region of subject noun, blue: region of object noun, yellow: the union bounding box).}
\label{fig:fig3}
\end{figure*}

\subsection{Visual Relationship between Objects in Images} \label{sec:SG}
\textbf{Semantic Object Relationship.}
We draw inspiration from recent advances in deep learning based visual relationship detection \cite{dai2017detecting,Yikang:ICCV17} and simplify it as a classification task to learn semantic relation classifier on visual relationship benchmarks (e.g., Visual Genome \cite{krishna2017visual}). The general expression of semantic relation is $\langle$\emph{subject-predicate-object}$\rangle$ between pairs of objects. Note that the semantic relation is directional, i.e., it relates one object (subject noun) and another object (object noun) via a predicate which can be an action or interaction between objects. Hence, given two detected regions of objects $v_i$ (subject noun) and $v_j$ (object noun) within an image ${I}$, we devise a simple deep classification model to predict the semantic relation between $v_i$ and $v_j$ depending on the union bounding box which covers the two objects together.

Figure \ref{fig:fig3} depicts the framework of our designed semantic relation detection model. In particular, the input two region-level features ${\bf{v}}_i$ and ${\bf{v}}_j$ are first separately transformed via an embedding layer, which are further concatenated with the transferred region-level feature ${\bf{v}}_{ij}$ of the union bounding box containing both $v_i$ and $v_j$. The combined features are finally injected into the classification layer that produces softmax probability over $N_{sem}$ semantic relation classes plus a non-relation class, which is essentially a multi-class logistic regression model. Here each region-level feature is taken from the $D_v$-dimensional ($D_v$=2,048) output of Pool5 layer after RoI pooling from the Res4b22 feature map of Faster R-CNN in conjunction with ResNet-101 \cite{He:CVPR16}.

After training the visual relation classifier on visual relationship benchmark, we directly employ the learnt visual relation classifier to construct the corresponding semantic graph $\mathcal{G}_{sem}=(\mathcal{V},\mathcal{E}_{sem})$. Specifically, we firstly group the detected $K$ image regions of objects within image ${I}$ into $K \times (K-1)$ object pairs (two identical regions will not be grouped). Next, we compute the probability distribution on all the $(N_{sem}+1)$ relation classes for each object pair with the learnt visual relation classifier. If the probability of non-relation class is less than 0.5, a directional edge from the region vertex of subject noun to the region vertex of object noun is established and the relation class with maximum probability is regarded as the label of this edge.

\begin{figure*}[!tb]
\centering {\includegraphics[width=0.9\textwidth]{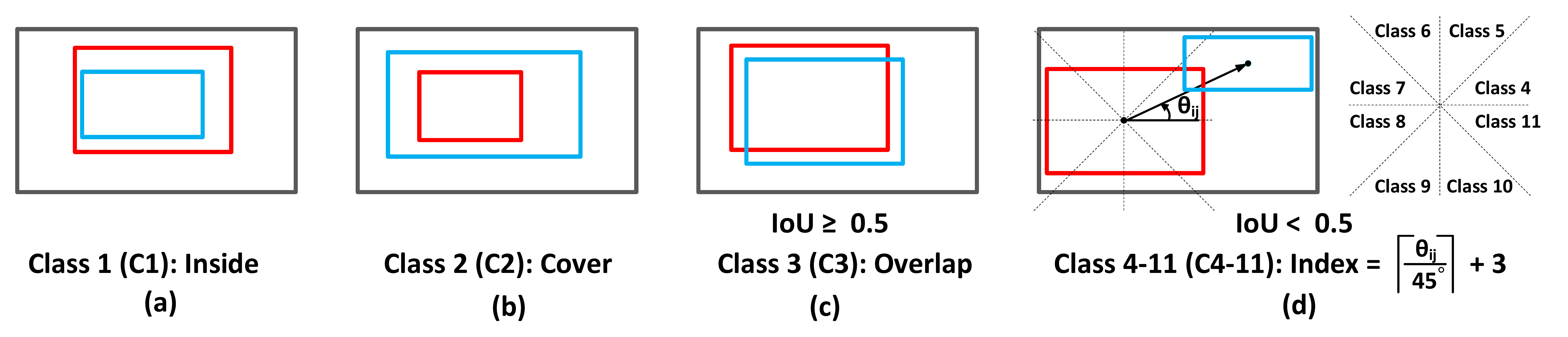}}
\caption{Definition of eleven kinds of spatial relations $\langle$\emph{object$_i$-object$_j$}$\rangle$ (red: region of object$_i$, blue: region of object$_j$).}
\label{fig:fig4}
\end{figure*}

\textbf{Spatial Object Relationship.}
The semantic graph only unfolds the inherent action/interaction between objects, while leaving the spatial relations between image regions unexploited. Therefore, we construct another graph, i.e., spatial graph, to fully explore the relative spatial relations between every two regions within one image. Here we generally express the directional spatial relation as $\langle$\emph{object$_i$-object$_j$}$\rangle$, which represents the relative geometrical position of object$_j$ against object$_i$. The edge and the corresponding class label for every two object vertices in spatial graph $\mathcal{G}_{spa}=(\mathcal{V},\mathcal{E}_{spa})$ are built and assigned depending on their Intersection over Union (IoU), relative distance and angle. Detailed definition of spatial relations are shown in Figure \ref{fig:fig4}.

Concretely, given two regions $v_i$ and $v_j$, the locations of them are denoted as $(x_i,y_i)$ and $(x_j,y_j)$, which are the normalized coordinates of the centroid of the bounding box on the image plane for $v_i$ and $v_j$, respectively. We can thus achieve the IoU between $v_i$ and $v_j$, relative distance ${d_{ij}}$ ({${d_{ij}} = \sqrt {{{\left( {{x_j} - {x_i}} \right)}^2} + {{\left( {{y_j} - {y_i}} \right)}^2}}$}) and relative angle ${\theta _{ij}}$ (i.e., the argument of the vector from the centroid of $v_i$ to that of $v_j$). Two kinds of special cases are firstly considered for classifying the spatial relation between $v_i$ and $v_j$. If $v_i$ completely includes $v_j$ or $v_i$ is fully covered by $v_j$, we establish an edge from $v_i$ to $v_j$ and set the label of spatial relation as ``inside" (\textbf{class 1}) and ``cover" (\textbf{class 2}), respectively. Except for the two special classes, if the IoU between $v_i$ and $v_j$ is larger than 0.5, we directly connect $v_i$ to $v_j$ with an edge, which is classified as ``overlap" (\textbf{class 3}). Otherwise, when the ratio $\phi_{ij}$ between the relative distance ${d_{ij}}$ and the diagonal length of the whole image is less than 0.5, we classify the edge between $v_i$ and $v_j$ solely relying on the size of relative angle ${\theta _{ij}}$ and the index of class is set as $\left\lceil {{{{\theta _{ij}}} \mathord{\left/{\vphantom {{{\theta _{ij}}} {{{45}^ \circ }}}} \right.\kern-\nulldelimiterspace} {{{45}^ \circ }}}} \right\rceil+3$ (\textbf{class 4-11}). When the ratio $\phi_{ij}>0.5$ and IoU $<0.5$, the spatial relation between them is tend to be weak and no edge is established in this case.

\subsection{Image Captioning with Visual Relationship}
With the constructed graphs over the detected objects based on their spatial and semantic connections, we next discuss how to integrate the learnt visual relationships into sequence learning with region-based attention mechanism for image captioning via our designed GCN-LSTM. Specifically, a GCN-based image encoder is devised to contextually encode all the image regions with semantic or spatial graph structure via GCN into relation-aware representations, which are further injected into attention LSTM for generating sentence.

\textbf{GCN-based Image Encoder.} Inspired from Graph Convolutional Networks for node classification \cite{kipf2017semi} and semantic role labeling \cite{marcheggiani2017encoding}, we design a GCN-based image encoder for enriching the region-level features by capturing the semantic/spatial relations on semantic/spatial graph, as illustrated in the middle part of Figure \ref{fig:fig2}. The original GCN is commonly operated on an undirected graph, encoding information about the neighborhood of each vertex $v_i$ as a real-valued vector, which is computed by
\begin{equation}\label{Eq:Eq2}
{\bf{v}}_i^{\left( 1 \right)} = \rho \big( {\sum\limits_{{v_j} \in \mathcal{N}( {{v_i}} )} {{\bf{W}}{{\bf{v}}_j}}  + {\bf{b}}} \big),
\end{equation}
where ${\bf{W}}\in {{\mathbb{R}}^{{D_v}\times {D_v}}}$ is the transformation matrix, ${\bf{b}}$ is the bias vector and $\rho$ denotes an activation function (e.g., ReLU). ${\mathcal{N}( {{v_i}} )}$ represents the set of neighbors of $v_i$, i.e., the region vertices have visual connections with $v_i$ here. Note that ${\mathcal{N}( {{v_i}} )}$ also includes $v_i$ itself. Although the original GCN refines each vertex by accumulating the features of its neighbors, none of the information about directionality or edge labels is included for encoding image regions. In order to enable the operation on labeled directional graph, the original GCN is upgraded by fully exploiting the directional and labeled visual connections between vertices.

Formally, consider a labeled directional graph $\mathcal{G}=(\mathcal{V},\mathcal{E})\in \{\mathcal{G}_{sem}, \mathcal{G}_{spa}\}$ where $\mathcal{V}$ is the set of all the detected region vertices and $\mathcal{E}$ is a set of visual relationship edges.
Separate transformation matrices and bias vectors are utilized for different directions and labels of edges, respectively, targeting for making the modified GCN sensitive to both directionality and labels. Accordingly, each vertex $v_i$ is encoded via the modified GCN as
\begin{equation}\label{Eq:Eq3}
{\bf{v}}_i^{\left( 1 \right)} = \rho \big( {\sum\limits_{{v_j} \in \mathcal{N}( {{v_i}} )} {{\bf{W}}_{\text{dir}(v_i,v_j)} {{\bf{v}}_j}}  + {\bf{b}}_{\text{lab}(v_i,v_j)}} \big),
\end{equation}
where ${\text{dir}(v_i,v_j)}$ selects the transformation matrix with regard to the directionality of each edge (i.e., ${\bf{W}}_1$ for $v_i$-to-$v_j$, ${\bf{W}}_2$ for $v_j$-to-$v_i$, and ${\bf{W}}_3$ for $v_i$-to-$v_i$). ${\text{lab}(v_i,v_j)}$ represents the label of each edge. Moreover, instead of uniformly accumulating the information from all connected vertices, an edge-wise gate unit is additionally incorporated into GCN to automatically focus on potentially important edges. Hence each vertex $v_i$ is finally encoded via the GCN in conjunction with an edge-wise gate as
\begin{equation}\label{Eq:Eq4}
\begin{array}{l}
{\bf{v}}_i^{\left( 1 \right)} = \rho \big( {\sum\limits_{{v_j} \in \mathcal{N}( {{v_i}} )} {g_{{v_i},{v_j}}}({{\bf{W}}_{\text{dir}(v_i,v_j)} {{\bf{v}}_j}}  + {\bf{b}}_{\text{lab}(v_i,v_j)})
} \big),\\
{g_{{v_i},{v_j}}} = \sigma \left( { {\widetilde{\bf{W}}_{\text{dir}(v_i,v_j)} {{\bf{v}}_j}}  + \widetilde{{b}}_{\text{lab}(v_i,v_j)} } \right),
\end{array}
\end{equation}
where ${g_{{v_i},{v_j}}}$ denotes the scale factor achieved from edge-wise gate, $\sigma$ is the logistic sigmoid function, $\widetilde{\bf{W}}_{\text{dir}(v_i,v_j)} \in {{\mathbb{R}}^ {1\times {D_v}}}$ is the transformation matrix and $\widetilde{{b}}_{\text{lab}(v_i,v_j)}\in {\mathbb{R}}$ is the bias. Consequently, after encoding all the regions $\{{{\bf{v}}_i}\}^{K}_{i=1}$ via GCN-based image encoder as in Eq.(\ref{Eq:Eq4}), the refined region-level features $\{{{\bf{v}}^{\left( 1 \right)}_i}\}^{K}_{i=1}$ are endowed with the inherent visual relationships between objects.

\textbf{Attention LSTM Sentence Decoder.}
Taking the inspiration from region-level attention mechanism in \cite{anderson2017bottom}, we devise our attention LSTM sentence decoder by injecting all of the relation-aware region-level features $\{{{\bf{v}}^{\left( 1 \right)}_i}\}^{K}_{i=1}$ into a two-layer LSTM with attention mechanism, as shown in the right part of Figure \ref{fig:fig2}. In particular, at each time step $t$, the attention LSTM decoder firstly collects the maximum contextual information by concatenating the input word $w_t$ with the previous output of the second-layer LSTM unit ${\bf{h}}^2_{t-1}$ and the mean-pooled image feature {$\overline {\bf{v}}  = \frac{1}{K}\sum\limits_{i = 1}^K {{\bf{v}}_i^{\left( 1 \right)}}$}, which will be set as the input of the first-layer LSTM unit. Hence the updating procedure for the first-layer LSTM unit is as
\begin{equation}\label{Eq:Eq5}
{\bf{h}}_t^1 = {f_1}\left( {\left[ {{\bf{h}}_{t - 1}^2,{{\bf{W}}_s}{{\bf{w}}_t},\overline {\bf{v}} } \right]} \right),
\end{equation}
where ${\bf{W}}_s\in {{\mathbb{R}}^{{D^1_s}\times {D_s}}}$ is the transformation matrix for input word $w_t$, ${\bf{h}}_t^1 \in {{\mathbb{R}}^{D_h}}$ is the output of the first-layer LSTM unit, and $f_1$ is the updating function within the first-layer LSTM unit. Next, depending on the output ${\bf{h}}_t^1$ of the first-layer LSTM unit, a normalized attention distribution over all the relation-aware region-level features is generated as
\begin{equation}\label{Eq:Eq6}
a_{t,i}={\bf{W}}_a\left[\tanh\left({\bf{W}}_f{{\bf{v}}_i^{\left( 1 \right)}} + {\bf{W}}_h{\bf{h}}_t^1\right)\right],~~~\lambda_t=softmax\left({\bf{a}}_t\right),
\end{equation}
where $a_{t,i}$ is the $i$-th element of ${\bf{a}}_t$, ${\bf{W}}_a\in {{\mathbb{R}}^{{1}\times {D_a}}}$, ${\bf{W}}_f\in {{\mathbb{R}}^{{D_a}\times {D_v}}}$ and ${\bf{W}}_h\in {{\mathbb{R}}^{{D_a}\times {D_h}}}$ are transformation matrices. $\lambda_t \in\mathbb R^{K}$ denotes the normalized attention distribution and its $i$-th element $\lambda_{t,i}$ is the attention probability of ${\bf{v}}_i^{\left( 1 \right)}$. Based on the attention distribution, we calculate the attended image feature {$\hat {\bf{v}}_t  = \sum\limits_{i = 1}^K {\lambda_{t,i}{\bf{v}}_i^{\left( 1 \right)}}$} by aggregating all the region-level features weighted with attention.  We further concatenate the attended image feature $\hat {\bf{v}}_t$ with ${\bf{h}}_t^1$ and feed them into the second-layer LSTM unit, whose updating procedure is thus given by
\begin{equation}\label{Eq:Eq7}
{\bf{h}}_t^2 = {f_2}\left( {\left[ {\hat {\bf{v}}_t,{\bf{h}}_{t}^1} \right]} \right),
\end{equation}
where $f_2$ is the updating function within the second-layer LSTM unit. The output of the second-layer LSTM unit ${\bf{h}}_t^2$ is leveraged to predict the next word $w_{t+1}$ through a softmax layer.

\subsection{Training and Inference}
In the training stage, we pre-construct the two kinds of visual graphs (i.e., semantic and spatial graphs) by exploiting the semantic and spatial relations among detected image regions as described in Section \ref{sec:SG}. Then, each graph is separately utilized to train one individual GCN-based encoder plus attention LSTM decoder. Note that the LSTM in decoder can be optimized with conventional cross entropy loss or the expected sentence-level reward loss as in \cite{Liu:2016PGSPIDEr,rennie2017self}.

At the inference time, we adopt a late fusion scheme to connect the two visual graphs in our designed GCN-LSTM architecture. Specifically, we linearly fuse the predicted word distributions from two decoders at each time step and pop out the word with the maximum probability as the input word to both decoders at the next time step. The fused probability for each word $w_i$ is calculated as:
\begin{equation}\label{Eq:Eq8}
\Pr \left( {{w_t} = {w_i}} \right) = \alpha {{\Pr}_{sem}}\left( {{w_t} = {w_i}} \right) + \left( {1 - \alpha } \right){{\Pr}_{spa}}\left( {{w_t} = {w_i}} \right),
\end{equation}
where $\alpha$ is the tradeoff parameter, ${{\Pr}_{sem}}\left( {{w_t} = {w_i}} \right)$ and ${{\Pr}_{spa}}\left( {{w_t} = {w_i}} \right)$ denotes the predicted probability for each word $w_i$ from the decoder trained with semantic and spatial graph, respectively.

\section{Experiments}
We conducted the experiments and evaluated our proposed GCN-LSTM model on COCO captioning dataset (COCO) \cite{Lin:ECCV14} for image captioning task. In addition, Visual Genome \cite{krishna2017visual} is utilized to pre-train the object detector and semantic relation detector in our GCN-LSTM.

\subsection{Datasets and Experimental Settings}
\textbf{COCO}, is the most popular benchmark for image captioning, which contains 82,783 training images and 40,504 validation images. There are 5 human-annotated descriptions per image. As the annotations of the official testing set are not publicly available, we follow the widely used settings in \cite{anderson2017bottom,rennie2017self} and take 113,287 images for training, 5K for validation and 5K for testing. Similar to \cite{Karpathy:CVPR15}, we convert all the descriptions in training set to lower case and discard rare words which occur less than 5 times, resulting in the final vocabulary with 10,201 unique words in COCO dataset.

\textbf{Visual Genome}, is a large-scale image dataset for modeling the interactions/relationships between objects, which contains 108K images with densely annotated objects, attributes, and relationships. To pre-train the object detector (i.e., Faster R-CNN in this work), we strictly follow the setting in \cite{anderson2017bottom}, taking 98K for training, 5K for validation and 5K for testing. Note that as part of images (about 51K) in Visual Genome are also found in COCO, the split of Visual Genome is carefully selected to avoid contamination of the COCO validation and testing sets. Similar to \cite{anderson2017bottom}, we perform extensive cleaning and filtering of training data, and train Faster R-CNN over the selected 1,600 object classes and 400 attributes classes. To pre-train the semantic relation detector, we adopt the same data split for training object detector. Moreover, we select the top-50 frequent predicates in training data and manually group them into 20 predicate/relation classes. The semantic relation detection model is thus trained over the 20 relation classes plus a non-relation class.

\textbf{Features and Parameter Settings.}
Each word in the sentence is represented as ``one-hot" vector (binary index vector in a vocabulary). For each image, we apply Faster R-CNN to detect objects within this image and select top $K=36$ regions with highest detection confidences to represent the image. Each region is represented as the 2,048-dimensional output of pool5 layer after RoI pooling from the Res4b22 feature map of Faster R-CNN in conjunction with ResNet-101 \cite{He:CVPR16}. In the attention LSTM decoder, the size of word embedding $D^1_s$ is set as 1,000. The dimension of the hidden layer $D_h$ in each LSTM is set as 1,000. The dimension of the hidden layer $D_a$ for measuring attention distribution is set as 512. The tradeoff parameter $\alpha$ in Eq.(\ref{Eq:Eq8}) is empirically set as $0.7$.

\textbf{Implementation Details.}
We mainly implement our GCN-LSTM based on Caffe \cite{Jia:MM14}, which is one of widely adopted deep learning frameworks. The whole system is trained by Adam \cite{kingma2014adam} optimizer. We set the initial learning rate as 0.0005 and the mini-batch size as 1,024. The maximum training iteration is set as 30K iterations. For sentence generation in inference stage, we adopt the beam search strategy and set the beam size as 3.

\textbf{Evaluation Metrics.} We adopt five types of metrics: BLEU@$N$ \cite{Papineni:ACL02}, METEOR \cite{Banerjee:ACL05}, ROUGE-L \cite{lin2004rouge}, CIDEr-D \cite{vedantam2015cider} and SPICE \cite{spice2016}. All the metrics are computed by using the codes\setcounter{footnote}{0}\footnote{\url {https://github.com/tylin/coco-caption}} released by COCO Evaluation Server \cite{chen2015microsoft}.

\textbf{Compared Approaches.}
We compared the following state-of-the-art methods: (1) \textbf{LSTM} \cite{Vinyals14} is the standard CNN plus RNN model which only injects image into LSTM at the initial time step. We directly extract results reported in \cite{rennie2017self}. (2) \textbf{SCST} \cite{rennie2017self} employs a modified visual attention mechanism of \cite{Xu:ICML15} for captioning. Moreover, a self-critical sequence training strategy is devised to train LSTM with expected sentence-level reward loss. (3) \textbf{ADP-ATT} \cite{Xiong2016MetaMind} develops an adaptive attention based encoder-decoder model for automatically determining when to look (sentinel gate) and where to look (spatial attention). (4) \textbf{LSTM-A} \cite{yao2017boosting} integrates semantic attributes into CNN plus RNN captioning model for boosting image captioning. (5) \textbf{Up-Down} \cite{anderson2017bottom} designs a combined bottom-up and top-down attention mechanism that enables region-level attention to be calculated. (6) \textbf{GCN-LSTM} is the proposal in this paper. Moreover, two slightly different settings of GCN-LSTM are named as GCN-LSTM$_{sem}$ and GCN-LSTM$_{spa}$ which are trained with only semantic graph and spatial graph,~respectively.

Note that for fair comparison, all the baselines and our model adopt ResNet-101 as the basic architecture of image feature extractor. Moreover, results are reported for models optimized with both cross entropy loss or expected sentence-level reward loss. The sentence-level reward is measured with CIDEr-D score.

\begin{table*}[t]
    \centering
    \setlength\tabcolsep{-0.5pt}
    \caption{Performance of our GCN-LSTM and other state-of-the-art methods on COCO, where B@$N$, M, R, C and S are short for BLEU@$N$, METEOR, ROUGE-L, CIDEr-D and SPICE scores. All values are reported as percentage (\%).}
    \begin{tabular}{l | c c c c c c | c c c c c c}
        \Xhline{2\arrayrulewidth}
		  & \multicolumn{6}{c|}{\textbf{Cross-Entropy Loss}} & \multicolumn{6}{c}{\textbf{CIDEr-D Score Optimization}} \\
		                  & B@1    & B@4    & M       & R      & C       & S      & B@1    & B@4    & M       & R      & C       & S    \\
	      \hline \hline
          LSTM \cite{Vinyals14}       & ~-~      & ~29.6~   & ~25.2~    & ~52.6~   & ~94.0~    & ~-~      & ~-~      & ~31.9~   & ~25.5~    & ~54.3~   & ~106.3~   & -    \\
          SCST \cite{rennie2017self}   & -      & 30.0   & 25.9    & 53.4   & 99.4    & -      & -      & 34.2   & 26.7    & 55.7   & 114.0   & -    \\
          ADP-ATT \cite{Xiong2016MetaMind}      & 74.2   & 33.2   & 26.6    & -      & 108.5   & -      & -      & -      & -       & -      & -       & -    \\
          LSTM-A \cite{yao2017boosting}       & 75.4   & 35.2   & 26.9    & 55.8   & 108.8   & 20.0   & 78.6   & 35.5   & 27.3    & 56.8   & 118.3   & 20.8 \\
          Up-Down \cite{anderson2017bottom}      & 77.2   & 36.2   & 27.0    & 56.4   & 113.5   & 20.3   & 79.8   & 36.3   & 27.7    & 56.9   & 120.1   & 21.4 \\\hline
          GCN-LSTM$_{spa}$ & 77.2   & 36.5   & 27.8    & 56.8   & 115.6   & 20.8   & 80.3   & 37.8   & 28.4    & 58.1   & 127.0   & 21.9 \\
          GCN-LSTM$_{sem}$ & 77.3   & 36.8   & 27.9    & 57.0   & 116.3   & 20.9   & 80.5   & 38.2   & 28.5    & 58.3   & 127.6   & 22.0 \\
          GCN-LSTM       & ~\textbf{77.4}~   & ~\textbf{37.1}~   & ~\textbf{28.1}~    & ~\textbf{57.2}~   & ~\textbf{117.1}~   & ~\textbf{21.1}~   & ~\textbf{80.9}~   & ~\textbf{38.3}~   & ~\textbf{28.6}~    & ~\textbf{58.5}~   & ~\textbf{128.7}~   & ~\textbf{22.1}~ \\
		\Xhline{2\arrayrulewidth}
    \end{tabular}
    \label{tab:COCO}
\end{table*}

\subsection{Performance Comparison and Experimental Analysis}
\textbf{Quantitative Analysis.} Table \ref{tab:COCO} shows the performances of different models on COCO image captioning dataset. Overall, the results across six evaluation metrics optimized with cross-entropy loss and CIDEr-D score consistently indicate that our proposed GCN-LSTM achieves superior performances against other state-of-the-art techniques including non-attention models (LSTM, LSTM-A) and attention-based approach (SCST, ADP-ATT and Up-Down). In particular, the CIDEr-D and SPICE scores of our GCN-LSTM can achieve 117.1\% and 21.1\% optimized with cross-entropy loss, making the relative improvement over the best competitor Up-Down by 3.2\% and 3.9\%, respectively, which is generally considered as a significant progress on this benchmark. As expected, the CIDEr-D and SPICE scores are boosted up to 128.7\% and 22.1\% when optimized with CIDEr-D score. LSTM-A exhibits better performance than LSTM, by further explicitly taking the high-level semantic information into account for encoding images. Moreover, SCST, ADP-ATT and Up-Down lead to a large performance boost over LSTM, which directly encodes image as one global representation. The results basically indicate the advantage of visual attention mechanism by learning to fucus on the image regions that are most indicative to infer the next word. More specifically, Up-Down by enabling attention to be calculated at the level of objects, improves SCST and ADP-ATT. The performances of Up-Down are still lower than our GCN-LSTM$_{spa}$ and GCN-LSTM$_{sem}$ which additionally exploits spatial/semantic relations between objects for enriching region-level representations and eventually enhancing image captioning, respectively. In addition, by utilizing both spatial and semantic graphs in a late fusion manner, our GCN-LSTM further boosts up the performances.

\begin{figure*}[!tb]
\centering {\includegraphics[width=0.9\textwidth]{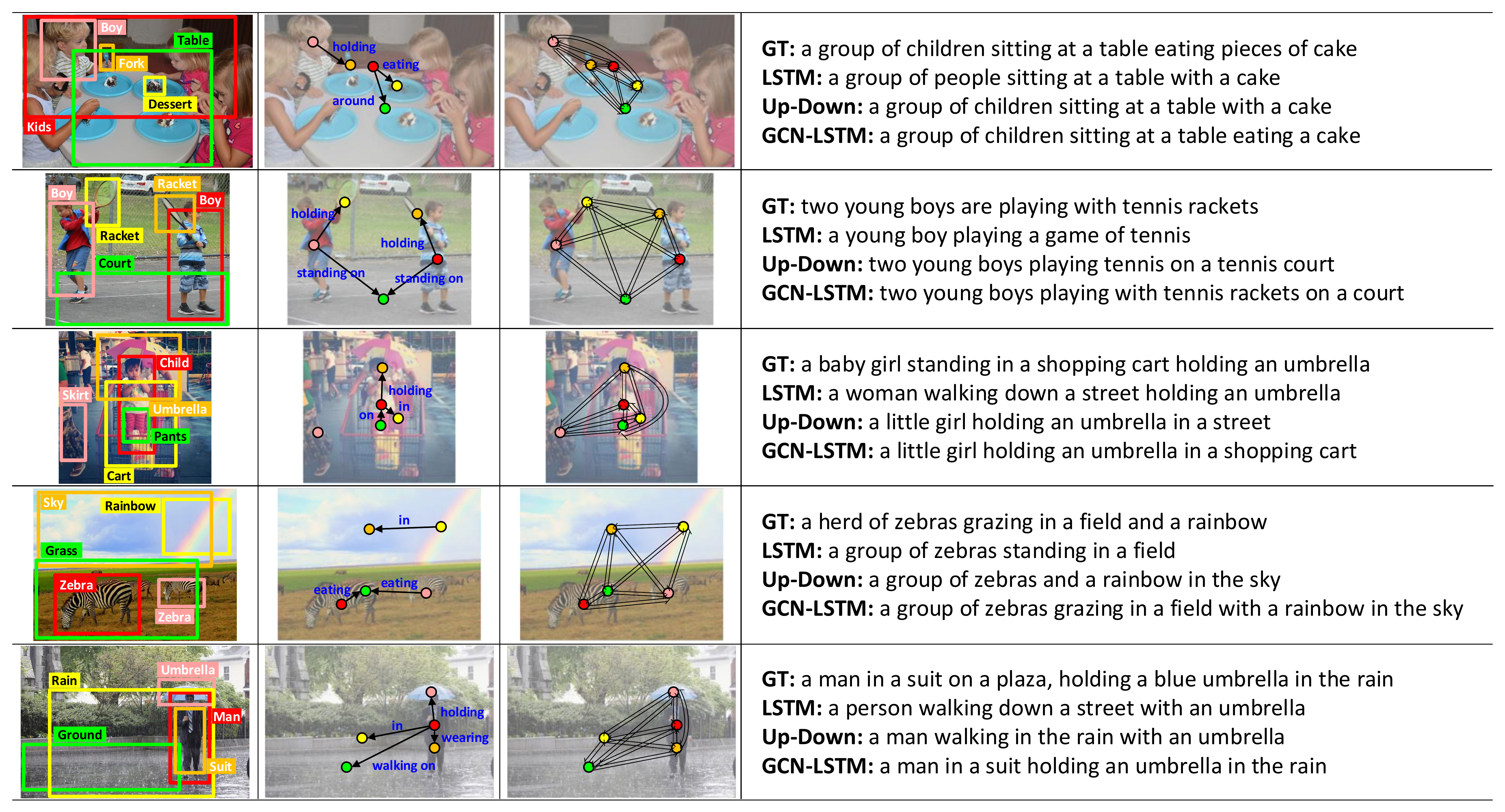}}
\caption{Graphs and sentences generation results on COCO dataset. The semantic graph is constructed with semantic relations predicted by our semantic relation detection model. The spatial graph is constructed with spatial relations as defined in Figure \ref{fig:fig4}. The output sentences are generated by 1) Ground Truth (GT): One ground truth sentence, 2) LSTM, 3) Up-Down and 4) our GCN-LSTM.}
\label{fig:figRS}
\end{figure*}

\textbf{Qualitative Analysis.} Figure \ref{fig:figRS} shows a few image examples with the constructed semantic and spatial graphs, human-annotated ground truth sentences and sentences generated by three approaches, i.e., LSTM, Up-Down and our GCN-LSTM. From these exemplar results, it is easy to see that the three automatic methods can generate somewhat relevant and logically correct sentences, while our model GCN-LSTM can generate more descriptive sentence by enriching semantics with visual relationships in graphs to boost image captioning. For instance, compared to the same sentence segment ``with a cake" in the sentences generated by LSTM and Up-Down for the first image, ``eating a cake" in our GCN-LSTM depicts the image content more comprehensive, since the detected relation ``eating" in semantic graph is encoded into relation-aware region-level features for guiding sentence generation.

\begin{table*}[!tb]
  \setlength\tabcolsep{2pt}
  \centering
  \caption{Leaderboard of the published state-of-the-art image captioning models on the online COCO testing server, where B@$N$, M, R, and C are short for BLEU@$N$, METEOR, ROUGE-L, and CIDEr-D scores. All values are reported as percentage (\%).}
  \label{table:leaderboard}

  \begin{tabular}{l|*{11}{c|}c}
  \Xhline{2\arrayrulewidth}
      \multicolumn{1}{c|}{\multirow{2}{*}{{Model}}} &  \multicolumn{2}{c|}{{B@2}} & \multicolumn{2}{c|}{{B@3}} & \multicolumn{2}{c|}{{B@4}} & \multicolumn{2}{c|}{{M}} & \multicolumn{2}{c|}{{R}} & \multicolumn{2}{c}{{C}} \\\cline{2-13}
      \multicolumn{1}{c|}{}&c5 &c40 &c5 &c40&c5 &c40&c5 &c40&c5 &c40&c5 &c40 \\\hline
      {GCN-LSTM} &  \textbf{65.5} &  \textbf{89.3} &  \textbf{50.8} &  \textbf{80.3} &  \textbf{38.7} &  \textbf{69.7} &  \textbf{28.5} &  \textbf{37.6} &  \textbf{58.5} &  \textbf{73.4} &  \textbf{125.3} &  \textbf{126.5} \\\hline
      {Up-Down} \cite{anderson2017bottom} &  64.1 &  88.8 &  49.1 &  79.4 &  36.9 &  68.5 &  27.6 &  36.7 &  57.1 &  72.4 &  117.9 &  120.5 \\\hline
      {LSTM-A} \cite{yao2017boosting} & 62.7 & 86.7 & 47.6  & 76.5 & 35.6 & 65.2 & 27.0 & 35.4 & 56.4 & 70.5 & 116.0 & 118.0 \\\hline
      {SCST} \cite{rennie2017self} &  61.9  &  86.0  &  47.0  &  75.9  &  35.2  &  64.5  & 27.0  &  35.5  &  56.3  &  70.7  &  114.7  &  116.7  \\\hline
      {G-RMI} \cite{Liu:2016PGSPIDEr}&  59.1  &  84.2  &  44.5  &  73.8  &  33.1  &  62.4  &  25.5  &  33.9  &  55.1  &  69.4  &  104.2  &  107.1 \\\hline
      {ADP-ATT} \cite{Xiong2016MetaMind}&  58.4  &  84.5  &  44.4  &  74.4  &  33.6  &  63.7  &  26.4  &  35.9  &  55.0  &  70.5  &  104.2  &  105.9 \\
      \Xhline{2\arrayrulewidth}
  \end{tabular}
\end{table*}

\textbf{Performance on COCO Online Testing Server.}
We also submitted our GCN-LSTM optimized with CIDEr-D score to online COCO testing server and evaluated the performance on official testing set. Table \ref{table:leaderboard} summarizes the performance Leaderboard on official testing image set with 5 (c5) and 40 (c40) reference captions. The latest top-5 performing methods which have been officially published are included in the table. Compared to the top performing methods on the leaderboard, our proposed GCN-LSTM achieves the best performances across all the evaluation metrics on both c5 and c40 testing sets.

\textbf{Human Evaluation.}
To better understand how satisfactory are the sentences generated from different methods, we also conducted a human study to compare our GCN-LSTM against two approaches, i.e., LSTM and Up-Down. All of the three methods are optimized with CIDEr-D score. 12 evaluators are invited and a subset of 1K images is randomly selected from testing set for the subjective evaluation. All the evaluators are organized into two groups. We show the first group all the three sentences generated by each approach plus five human-annotated sentences and ask them the question: Do the systems produce captions resembling human-generated sentences? In contrast, we show the second group once only one sentence generated by different approach or human annotation (Human) and they are asked: Can you determine whether the given sentence has been generated by a system or by a human being? From evaluators' responses, we calculate two metrics: 1) M1: percentage of captions that are evaluated as better or equal to human caption; 2) M2: percentage of captions that pass the Turing Test. The results of M1 are 74.2\%, 70.3\%, 50.1\% for GCN-LSTM, Up-Down and LSTM. For the M2 metric, the results of Human, GCN-LSTM, Up-Down and LSTM are 92.6\%, 82.1\%, 78.5\% and 57.8\%. Overall, our GCN-LSTM is clearly the winner in terms of two criteria.

\begin{figure*}[!tb]
\centering {\includegraphics[width=0.8\textwidth]{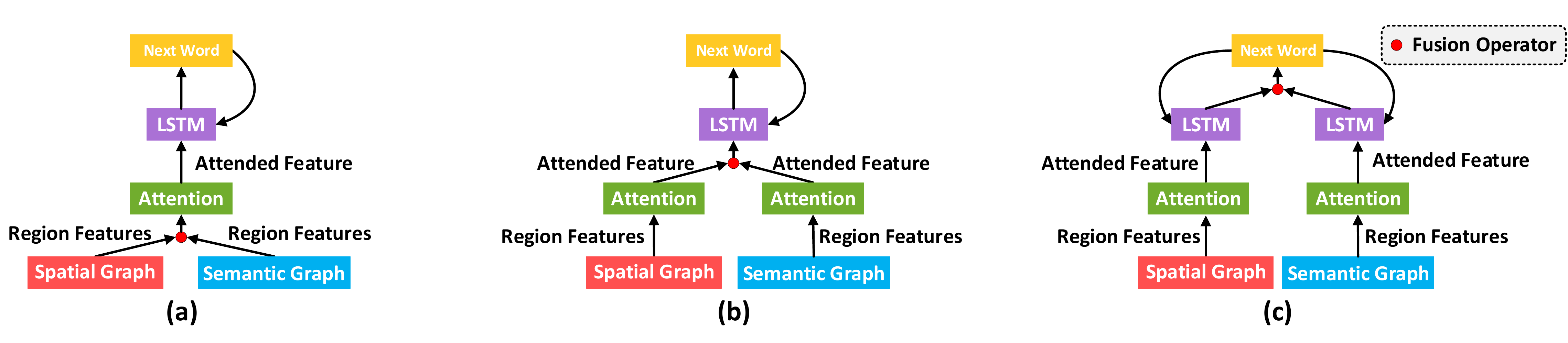}}
\caption{Different schemes for fusing spatial and semantic graphs in GCN-LSTM: (a) Early fusion before attention module, (b) Early fusion after attention module and (c) Late fusion. The fusion operator could be concatenation or summation.}
\label{fig:figFusion}
\end{figure*}

\textbf{Effect of Fusion Scheme.} There are generally two directions for fusing semantic and spatial graphs in GCN-LSTM. One is to perform early fusion scheme by concatenating each pair of region features from graphs before attention module or the attended features from graphs after attention module. The other is our adopted late fusion scheme to linearly fuse the predicted word distributions from two decoders. Figure \ref{fig:figFusion} depicts the three fusion schemes. We compare the performances of our GCN-LSTM in the three fusion schemes (with cross-entropy loss). The results are 116.4\%, 116.6\% and 117.1\% in CIDEr-D metric for early fusion before/after attention module and late fusion, respectively, which indicate that the adopted late fusion scheme outperforms other two early fusion schemes.

\begin{figure*}[!tb]
\centering {\includegraphics[width=0.7\textwidth]{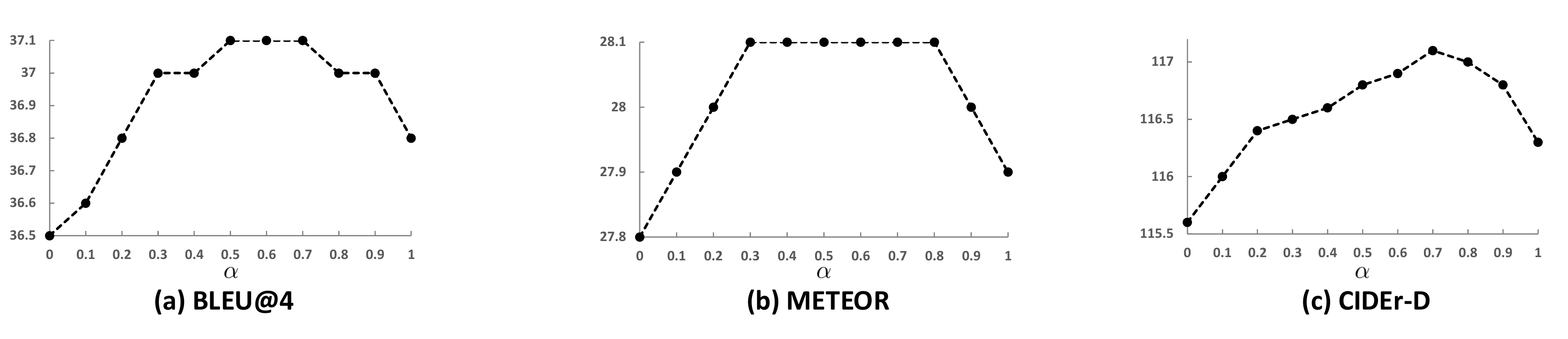}}
\caption{The effect of the tradeoff parameter $\alpha$ in our GCN-LSTM with cross-entropy loss over (a) BLEU@4 (\%), (b) METEOR (\%) and (c) CIDEr-D (\%) on COCO.}
\label{fig:figtradeoff}
\end{figure*}

\textbf{Effect of the Tradeoff Parameter $\alpha$.}
To clarify the effect of the tradeoff parameter $\alpha$ in Eq.(\ref{Eq:Eq8}), we illustrate the performance curves over three evaluation metrics with a different tradeoff parameter in Figure \ref{fig:figtradeoff}. As shown in the figure, we can see that all performance curves are generally like the ``$\wedge$" shapes when $\alpha$ varies in a range from 0 to 1. The best performance is achieved when $\alpha$ is about 0.7. This proves that it is reasonable to exploit both semantic and spatial relations between objects for boosting image captioning.

\section{Conclusions}
We have presented Graph Convolutional Networks plus Long Short-Term Memory (GCN-LSTM) architecture, which explores visual relationship for boosting image captioning. Particularly, we study the problem from the viewpoint of modeling mutual interactions between objects/regions to enrich region-level representations that are feed into sentence decoder. To verify our claim, we have built two kinds of visual relationships, i.e., semantic and spatial correlations, on the detected regions, and devised Graph Convolutions on the region-level representations with visual relationships to learn more powerful representations. Such relation-aware region-level representations are then input into attention LSTM for sentence generation. Extensive experiments conducted on COCO image captioning dataset validate our proposal and analysis. More remarkably, we achieve new state-of-the-art performances on this dataset. One possible future direction would be to generalize relationship modeling and utilization to other vision tasks.

\bibliographystyle{splncs04}
\bibliography{egbib}

\end{document}